# Learning based Predictive Error Estimation and Compensator Design for Autonomous Vehicle Path Tracking

Chaoyang Jiang, Hanqing Tian, Jibin Hu, Jiankun Zhai, Chao Wei, and Jun Ni

*Abstract*—Model predictive control (MPC) is widely used for path tracking of autonomous vehicles due to its ability to handle various types of constraints. However, a considerable predictive error exists because of the error of mathematics model or the model linearization. In this paper, we propose a framework combining the MPC with a learning-based error estimator and a feedforward compensator to improve the path tracking accuracy. An extreme learning machine is implemented to estimate the model based predictive error from vehicle state feedback information. Offline training data is collected from a vehicle controlled by a model-defective regular MPC for path tracking in several working conditions, respectively. The data include vehicle state and the spatial error between the current actual position and the corresponding predictive position. According to the estimated predictive error, we then design a PID-based feedforward compensator. Simulation results via Carsim show the estimation accuracy of the predictive error and the effectiveness of the proposed framework for path tracking of an autonomous vehicle.

## I. INTRODUCTION

THE path tracking of autonomous vehicles has attracted attentions in the past decade. In some early research works, the autonomous vehicle was modeled as wheeled robot by kinematic equation with nonholonomic constraints. The path following controller focused on the kinematic model and had a good performance in low speed conditions. The pure-pursuit and Stanley method were popular since The DARPA Challenge competition. Thanks to the development of hardware and the improvement of computational resources, the model predictive control (MPC) became more popular in path tracking mission of both wheeled robots and high-speed automobiles due to the ability of straightforward handling nonlinear dynamic system and multiple constraints [1-3].

Theoretically, with the real dynamic model of an autonomous vehicle, MPC can handle the path following task with preferable stability and accuracy [4]. However, it is a great challenge to identity the real dynamic model due to the system nonlinearity and parameter uncertainty. Additionally, because of the computational effort limitation, it is difficult to online solve an iteratively nonlinear programming problem. Hence, the predictive model usually has to be simplified by several kinds of approximations [5]. Therefore, it can be inferred that the performance of MPC suffers from a considerable predictive error caused by system model simplification and linearization or the inherent modeling error.

Some recent works provided the modifications using learning methods or adaptive methods to enhance the traditional MPC and improve control performance in terms of economic efficiency [6], time consumption in iteration tasks [7], vehicle speed [8] and robustness to system offset [9], etc. Minimizing the modeling error via learning methods is one of the effective steps for the control performance enhancement. There are also recent researches about learning driver model and personalized tracking control [10].

In this paper, we propose an original framework in which the predictive error is estimated via a machine learning technique and a feedforward component is introduced based on the learned predictive error to compensate the traditional MPC. The predictive error is estimated by a single layer feedforward neural network that is trained by the Extreme Learning Machine (ELM) technique, which has good capability in nonlinear regression task or learning a representation from offline training data. The ELM is computationally efficient and shown effective in estimating the predictive error involved in a MPC of an autonomous vehicle. The proposed framework

This work was supported by the National Natural Science Foundation of China under Grant 51875039 and the Beijing Institute of Technology Research Fund Program for Young Scholars.

The authors are with the National Key Lab of Vehicular Transmission, School of Mechanical Engineering, Beijing Institute of Technology, Beijing, China. (cjiang@bit.edu.cn, 13512259231@163.com)

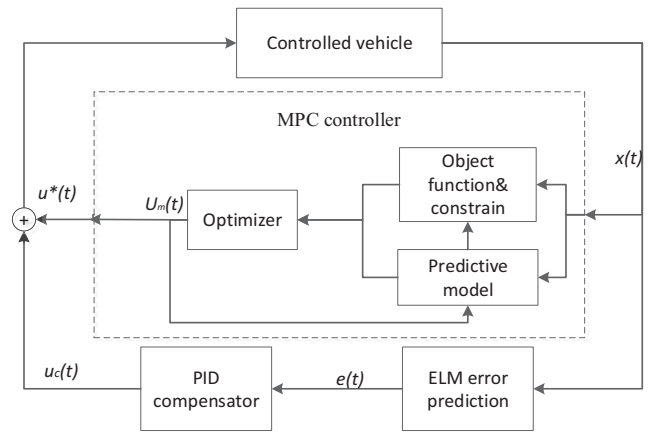

Fig. 1. Proposed system framework.

combines the data-driving technique with the traditional MPC. With the estimated predictive error, a compensator is designed to enhance the vehicle control input and improve the path tracking performance.

The rest of the paper is organized as follows: the overall structure of the proposed framework is shown in section II. A description of modeling and MPC for path tracking is given in section III followed by the ELM based error estimation and compensator design in section IV. The simulation design and validation results are presented in section V. The paper is concluded in the section VI. Finally, an appendix for the parameters list used in the paper is provided.

## II. OVERALL STRUCTURE OF THE PROPOSED FRAMEWORK

As shown in Fig.1, the standard MPC is extended and modified by adding a feedforward compensator, which is designed from the estimated predictive error.

### A. Model predictive control for path tracking

The structure of the MPC for path tracking shows in the dashed box in Fig. 1. The main processes include model prediction, online rolling optimization and error feedback. Given a optimization object function, the optimizer can find the control input by solving a quadratic programming problem online. The controller works independently to other parts of the system as a feedback controller and contributes the main part of the control input of the vehicle. However, the gap between the model and the real system lead to a considerable predictive error and affects the control performance.

### B. Learning based Error Estimation

An extreme learning machine (ELM) is implemented to estimate the model predictive error of MPC, that is, the difference between the real and the predictive trajectory of each computation iteration of MPC. As shown in Fig.2, although MPC uses feedback to iteratively renew the prediction in the predictive horizon in order to eliminate accumulative error, there is always a single-step predictive error to the real trajectory.

The dataset is collected from a certain vehicle plant in Carsim controlled by a standard MPC under several working conditions, which include double lane change, constant turn and straight line. The vehicle follows the path with constant longitudinal speed. The input of the ELM is an eight-dimensional vector of vehicle state signals including longitudinal and lateral position $X$ and $Y$, yaw heading $\varphi$, yaw rate $\omega$, longitudinal and lateral velocity $V_x$ and $V_y$, slip ratio of the two front wheels. The expected output is the predictive error $e(t)$ between predictive position $(X_p, Y_p)$ and the real vehicle position $(X, Y)$ of each predictive iteration. With the estimated error, a feedforward compensator can be designed to enhance the tracking performance.

### C. Feedforward Compensator

We implement PID to design the compensator since it can work without system model and its parameters can be tuned by experiment, that is, the feedforward $u_c$ is calculated by a PID function of the estimated predictive error. The final control input of the vehicle is contributed by the output of the original MPC and the compensator.

## III. VEHICLE DYNAMIC MODELING AND MPC FOR PATH TRACKING

As a non-linear system with many kinematic and dynamic constrains, full-scale vehicle has a more complicated dynamic model than wheeled robots. Considering the real-time computation ability and computational resources consuming, the predictive model must be simplified. For path tracking control, we focus on the longitudinal and lateral dynamic of the vehicle so the three-degree-of-freedom bicycle model is widely used [11-12].

### A. Vehicle Bicycle Model with small steering angle and linear tire model

Nonlinear three-DOF vehicle dynamic model is still a complicated model for predictive control. In addition, the nonlinear relationship between tire force and tire slip angle also increases the computational difficulty. In the previous research, the experimental analysis shows that linear function can approximately represent the conventional tire model when the lateral acceleration $a_y \leq 0.4g$

$$F_l = C_l s \quad F_c = C_c \alpha \tag{1}$$

where $C_l$ is the tire longitudinal stiffness and $C_c$ is the lateral stiffness.

Under the small steering angle and linear tire model assumption, we can obtain

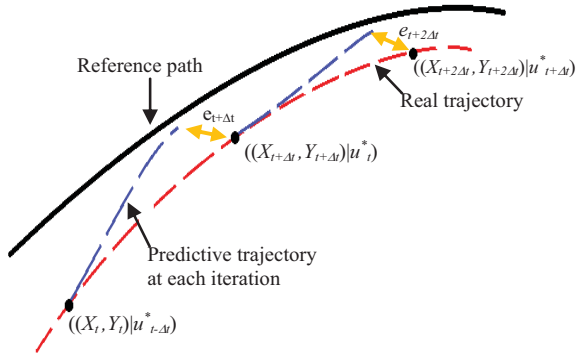

Fig.2 Predictive error caused by model error

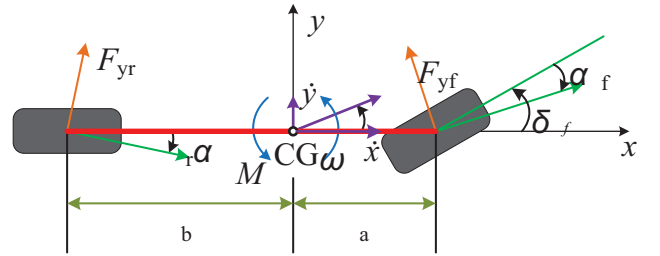

Fig. 3. Vehicle bicycle model

$$m\ddot{y} = -m\dot{x}\dot{\varphi} + 2\left[C_{cf}\left(\delta_f - \frac{\dot{y}+a\dot{\varphi}}{\dot{x}}\right) + C_{cr}\frac{b\dot{\varphi}-\dot{y}}{\dot{x}}\right] \quad (2)$$

$$m\ddot{x} = m\dot{y}\dot{\varphi} + 2\left[C_{lf}s_f + C_{cf}\left(\delta_f - \frac{\dot{y}+a\dot{\varphi}}{\dot{x}}\right)\delta_f + C_{lr}s_r\right]$$

$$I_z\ddot{\varphi} = 2\left[aC_{cf}\left(\delta_f - \frac{\dot{y}+a\dot{\varphi}}{\dot{x}}\right) - bC_{cr}\frac{b\dot{\varphi}-\dot{y}}{\dot{x}}\right]$$

$$\dot{Y} = \dot{x}\sin\varphi + \dot{y}\cos\varphi$$

$$\dot{X} = \dot{x}\cos\varphi - \dot{y}\sin\varphi$$

Here, we set $x_{d,v} = [\dot{y}\ \dot{x}\ \varphi\ \dot{\varphi}\ Y\ X]^T$ as the state variables and $u = \delta_f$ as the control input. All parameters are listed in Table I as shown in the appendix.

### B. Model Predictive Error Analysis

The bicycle model is formulated under some simplifications and assumptions. Firstly, we assume that the linear function can fit the tire model. This assumption is tenable when tire force and sideslip angle are in a low range but often fails when vehicle has a large lateral force during extreme handling conditions. The nonlinear error caused by simplified modeling becomes not acceptable. The tire nonlinear dynamics is mainly responsible for the model error.

Secondly, it is assumed that the front wheel steering angle and tire sideslip angle are small enough, and the trigonometric functions can be approximated

$$\sin\theta \approx \theta, \cos\theta \approx 1 \quad (3)$$

where $\theta$ represents both the front wheel steering angle and tire sideslip angle. Error may show during a large curvature bend.

Thirdly, the vehicle parameters including the position of gravity center (GC), the tire pressure, the wheel track and the mass of the whole vehicle are hard to measure precisely in practice. Some of them may change under different working conditions. The inaccuracy in modeling parameters leads to predictive error. There are even time-variant errors caused by mechanical systems like bias or lost motion of steering and the motion coupling of the suspension system.

### C. MPC for Path Tracking

A MPC for path tracking using the similar linearization technique proposed in [3] and bicycle model is implemented. By keeping a constant vehicle speed, the system can be simplified to a 2-DOF system with a single input $u = \delta_f$ since we mainly focus on lateral control. Therefore, the state-space equations are built.

$$\xi(k+1|t) = \tilde{A}_{k,t}\xi(k|t) + \tilde{B}_{k,t}\Delta u(k|t) \quad (4)$$

$$y(k|t) = \tilde{C}_{k,t}\xi(k|t)$$

where $\xi(k|t) = \begin{bmatrix} x(k|t) \\ u(k-1|t) \end{bmatrix}$, $\tilde{A}_{k,t} = \begin{bmatrix} A_{k,t} & B_{k,t} \\ 0_{m\times n} & I_m \end{bmatrix}$,

$\tilde{B}_{k,t} = \begin{bmatrix} B_{k,t} \\ I_m \end{bmatrix}$, $\tilde{C}_{k,t} = \begin{bmatrix} C_{k,t} & 0 \end{bmatrix}$. $A_{k,t}$ and $B_{k,t}$ are solved by following linearization and then discretization from the vehicle model equation (2) shown in section A.

$$A_{k,t} = I + \Delta T A_{t,t}, B_{k,t} = \Delta T B_{t,t}$$

where $\Delta T$ is the sample time interval of the controller. $A_{t,t}, B_{t,t}$ are the jacobian matrixes about $x_t$ and $u_t$ for the nonlinear system (2) at $t$, respectively.

We choose the optimization function with the similar form in [3] by considering the control accuracy and effort then transfer the solving process into a quadratic programming

$$J(\xi(t),u(t-1),\Delta U(t)) = \sum_{i=1}^{N_p}\left\|y(t+i|t) - y_{ref}(t+i|t)\right\|_Q^2 +$$

$$\sum_{i=1}^{N_c-1}\|\Delta u(t+i|t)\|_R^2 \quad (5)$$

$$s.t.\begin{cases} \Delta U_{\min} \leq \Delta U_{k,t} \leq \Delta U_{\max} \\ U_{\min} \leq \Delta U_{k,t} + U_t \leq U_{\max} \end{cases}$$

where $y$ is the predictive state in (4). $\Delta U_{\max}, \Delta U_{\min}, U_{\max}, U_{\min}$ are the control increment and input boundaries.

After solving the QP problem in each control iteration, we can get the optimal control increments sequence and apply the current control output $u_m(t)$

$$\Delta U_t^* = \left[\Delta u_t^*, \Delta u_{t+1}^*, \cdots, \Delta u_{t+N_c-1}^*\right] \quad (6)$$

$$u_m(t) = u_m(t-1) + \Delta u_t^*$$

It is worth mentioning that the vehicle model implemented in the MPC has small discrepancy with the setting in Carsim in order to make obvious predictive error and to verify the effectiveness of the feedforward compensator. We add predictive inaccuracy by changing parameters of the vehicle model such as GC position and tire type, which will change the property of the system to some extent.

## IV. ELM BASED PREDICTIVE ERROR ESTIMATION AND COMPENSATOR DESIGN

### A. ELM for Predictive Error Estimation

It has been verified that extreme learning machine has a good performance in regression and classification task on small scale dataset with a relatively fast training speed than other learning methods such as BP neural network or SVM, and hence has been successfully applied in many applications [13-15]. ELM can obtain higher overall accuracy and greatly reduce training time because the input weight matrix and hidden layer bias can be randomly generated without training.

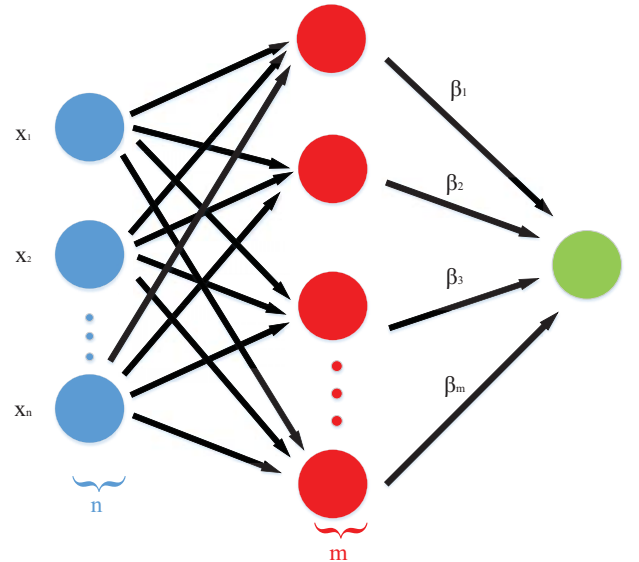

Fig. 4. Extreme learning machine structure.

In this work, the data set is collected from simulation which contains 1250 samples, and 250 of them are randomly selected for testing, which are the typical small-scale dataset. Therefore, we eventually select ELM as the data-driven technique for predictive error estimation. The result is shown later.

The predictive error estimation is a typical regression process by taking vehicle state $\xi_{d,v}$ as input. The network structure of ELM is shown in Fig. 4. The input layer has 8 nodes. With well-tuning based on the dataset, we set the single hidden layer 55 nodes. In an ELM with single hidden layer, we define the output equation of the hidden node $i$ as

$$h_i(x) = G(a_i \cdot x + b_i) \quad (7)$$

where $a_i$ and $b_i$ are the parameters of the hidden node and $a_i$ is the input weight. $G$ is the activation function.

The regression output of an ELM with $l$ hidden nodes is

$$f_L(x) = \sum_{i=1}^{l} \beta_i h_i(x) \quad (8)$$

where $\beta_i$ is the output weight of the hidden node.

The training process is to solve output weights with the given objective function defined as

$$J = \|\beta\|_p^{\sigma_1} + C\|H\beta - L\|_q^{\sigma_2} \quad (9)$$

where $L$ is target labels, $C$ is regularization coefficient and $H$ is the hidden layer output matrix. Then the output weight parameter can be calculated as

$$\beta = \left(\frac{I}{C} + H^T H\right)^{-1} H^T L \quad (10)$$

### B. Compensator Design

After estimating the error, we design a compensator using proportional control since it does not need a system model and the parameter can be adjusted easily

$$u_c = K_p e \quad (11)$$

Here $e$ is the estimated predictive error. The final control input applied to the vehicle is the summation of the MPC output and compensator output, that is

$$u^* = u_m + u_c. \quad (12)$$

## V. SIMULATION

To validate the proposed framework, we apply the Simulink and Carsim co-simulation platform to accomplish a path tracking test. The reference path is designed as a non-equidistant double-lane changing determined by nonlinear function.

We mainly focus on testing the lateral error of the path following and the vehicle test speed is set as a constant in Carsim. The vehicle keeps 75km/h in the simulation and has a large lateral acceleration during turning. In this condition, the tire model is obviously nonlinear so the error can be significant. Therefore, the test run speed is carried out in a high-speed working condition to prove the effectiveness of proposed method in improving control performance.

Firstly, we test the MPC-only controller with small modeling parameters discrepancy as the baseline. We well-tune the quadratic weight matrixes Q and R in the objective function.

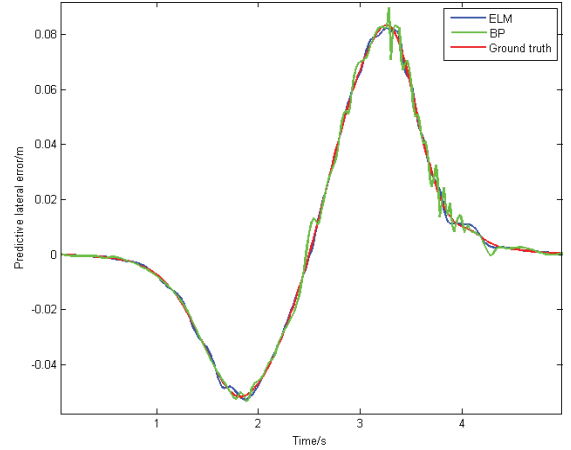

Fig. 5. The predictive error estimation

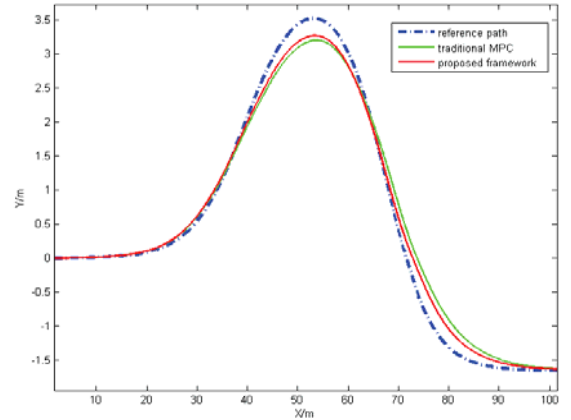

Fig. 6. The path tracking trajectories

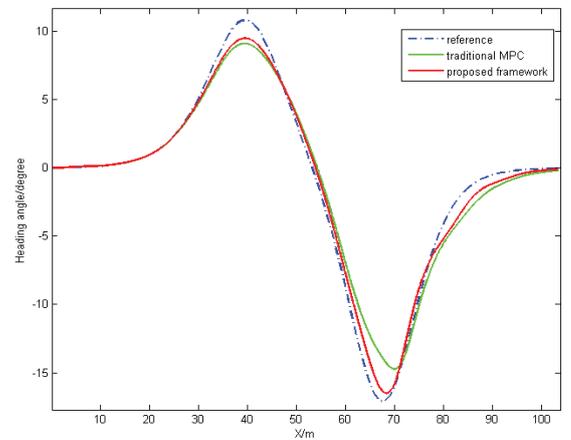

Fig. 7. The heading angles

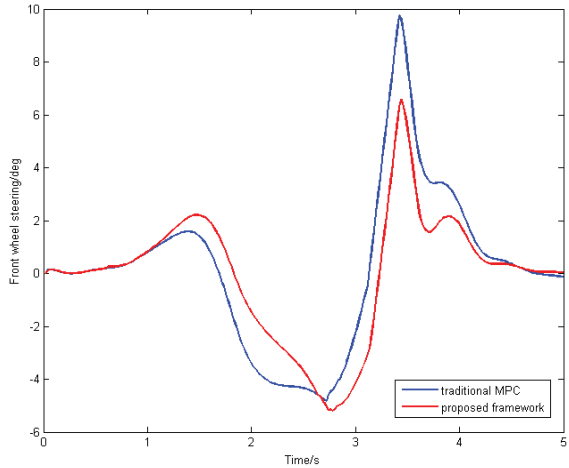

Fig. 8. The vehicle control input

The vehicle can accomplish a smooth and stable path tracking task with lateral error occurred mainly at large steering condition.

Then, we integrate an offline-trained ELM as error estimator with a PID compensator. Before online closed-loop simulation test, we validate the ability of predictive error estimation. The test data is collected during the prior MPC-only test. The error estimation results are shown in Fig.5. In addition, the performance comparison with BP as a baseline shows that the results of the ELM can well-fit the target label and outperform that of the BP.

Finally, we conduct the online testing of the proposed framework. Fig.6 shows the result of the two trajectories. There is a significant improvement in tracking accuracy after adding the feedforward compensator. At the position that X=80m, the lateral error can be reduced by 0.15m which equals to nearly 30% of the total lateral error. Fig.7 shows the heading angle to the reference path. The proposed method also outperforms the MPC-only method. It has to be mentioned that the compensator can only eliminate the error caused by the model error. In another word, since the total tracking error is also caused by the balance between tracking accuracy, control energy and safety constraint, it cannot be eliminated.

Fig.8 shows the control input $u^*(t)$ of the two methods. The MPC-only method shows a larger control cost for the same reference tracking task. The proposed method reduces the maximum steering angle, which makes the whole driving process more smoothing and stable.

## VI. Concluding Remarks and Future works

In this paper, we provide a framework in which the predictive error involved in an MPC is estimated via an ELM and is applied to design a feedforward compensator of the MPC. The compensator is simply designed as a proportional component in terms of the estimated predictive error. Simulation results show that the proposed framework can improve the path tracking performance of autonomous vehicles compared with the traditional MPC. The future work will focus on two main directions: the experiments on a full-size vehicle as validation and the refined design of the compensator.

## Appendix

TABLE I
PARAMETERS LIST

| Parameters | Value |
|---|---|
| $(x, y)$ | Center of gravity in global coordinates |
| $\varphi$ | Vehicle's heading angle |
| $\dot{x}$ | Longitudinal speed |
| $\dot{y}$ | Lateral speed |
| $\dot{\varphi}$ | Yaw rate |
| $\delta_f$ | *Steering angle* |
| $m$ | Vehicle's mass |
| $I_z$ | Moment of inertia |
| $a$ | Distances between the vehicles center of gravity and the front axle |
| $b$ | Distances between the vehicles center of gravity and the rear axle |
| $s$ | Slip ratio |
| $C_{lf}$ | Front tire longitudinal stiffness |
| $C_{lr}$ | Rear tire longitudinal stiffness |
| $C_{cf}$ | Front tire lateral stiffness |
| $C_{cr}$ | Rear tire lateral stiffness |

## References

[1] O. Pauca, C. F. Caruntu and C. Lazar, "Predictive Control for the Lateral and Longitudinal Dynamics in Automated Vehicles," 2019 23rd International Conference on System Theory, Control and Computing (ICSTCC), Sinaia, Romania, 2019, pp. 797-802.
[2] Y. Xu, B. Chen, X. Shan, W. Jia, Z. Lu and G. Xu, "Model predictive control for lane keeping system in autonomous vehicle", Proc. 7th Int. Conf. Power Electron. Syst. Appl.-Smart Mobility Power Transfer Secur., pp. 1-5, 2017.
[3] F. Kuhne, W. F. Lages, and JG Da Silva Jr, Model Predictive Control of a Mobile Robot Using Linearization. Proceedings of Mechatronic and Robotics, 525-530, 2004.
[4] J. Kong, M. Pfeiffer, G. Schildbach and F. Borrelli, "Kinematic and dynamic vehicle models for autonomous driving control design," 2015 IEEE Intelligent Vehicles Symposium (IV), Seoul, 2015, pp. 1094-1099.
[5] Z. Ercan, M. Gokasan and F. Borrelli, "An adaptive and predictive controller design for lateral control of an autonomous vehicle," 2017 IEEE International Conference on Vehicular Electronics and Safety (ICVES), Vienna, 2017, pp. 13-18.
[6] B. Sakhdari and N. Azad, Adaptive Tube-based Nonlinear MPC for Economic Autonomous Cruise Control of Plug-in Hybrid Electric Vehicles. IEEE Transactions on Vehicular Technology. 67(12):11390-11401, 2018.
[7] U. Rosolia and F. Borrelli, Learning Model Predictive Control for iterative tasks. A Data-Driven Control Framework, IEEE Transactions on Automatic Control, 63(7):1883-1896, 2017.
[8] G. Williams et al., Locally Weighted Regression Pseudo-Rehearsal for Online Learning of Vehicle Dynamics. arXiv preprint arXiv:1905.05162 , 2019.
[9] M. Bujarbaruah et al. Adaptive MPC for Autonomous Lane Keeping. arXiv preprint arXiv:1806.04335, 2018
[10] Z. Li, B. Wang, J. Gong, T. Gao, C. Lu, and G. Wang, ''Development and evaluation of two learning-based personalized driver models for pure pursuit path-tracking behaviors,'' in Proc. IEEE Intell. Vehicles Symp. (IV), Jun. 2018, pp. 79–84.
[11] H. Pacejka, Tyre and Vehicle Dynamics (Second edition), Butterworth-Heinemann, 2005.
[12] N. Zhang, J. Ni, and J. Hu, Robust H∞ state feedback control for handling stability of intelligent vehicles on a novel all-wheel independent steering mode. IET Intelligent Transport Systems, 13(10): 1579-1589,2019.
[13] G. B. Huang, Q. Y. Zhu, and C. K. Siew, Extreme learning machine: theory and applications. Neurocomputing, 70(1-3): 489-501, 2006.
[14] C. Jiang, M. K. Masood, Y. C. Soh, and H. Li, Indoor occupancy estimation from carbon dioxide concentration, Energy and Buildings, 131:132-141, 2016.
[15] Z. Chen, C. Jiang, and L. Xie, A novel ensemble ELM for human activity recognition using smartphone sensors, IEEE Transactions on Industrial Informatics, 15(5):2691-2699, 2019.